\documentclass[10pt,twocolumn,letterpaper]{article}

\usepackage{cvpr}
\usepackage{times}
\usepackage{epsfig}
\usepackage{graphicx}
\usepackage{amsmath}
\usepackage{amssymb}
\usepackage[utf8]{inputenc}
\usepackage[table]{xcolor}
\usepackage{makecell}
\usepackage{overpic}
\usepackage{subcaption}
\usepackage{slashbox}
\usepackage{multirow}
\usepackage{gensymb}
\usepackage{hhline}

\usepackage[pagebackref=true,breaklinks=true,letterpaper=true,colorlinks,bookmarks=false]{hyperref}

\usepackage[sort&compress,capitalize,noabbrev,nameinlink]{cleveref}
\creflabelformat{equation}{#2#1#3}
\crefformat{footnote}{#2\footnotemark[#1]#3}

\newcommand\Tstrut{\rule{0pt}{2.6ex}}         
\newcommand\Bstrut{\rule[-0.8ex]{0pt}{0pt}}   

\cvprfinalcopy 

\def\cvprPaperID{2} 

\ifcvprfinal\fi
\begin{document}

\title{An Empirical Evaluation Study on the Training of\\SDC Features for Dense Pixel Matching}

\author{René Schuster\textsuperscript{1} \hspace{0.5cm} Oliver Wasenmüller\textsuperscript{1} \hspace{0.5cm} Christian Unger\textsuperscript{2} \hspace{0.5cm} Didier Stricker\textsuperscript{1}\\
\textsuperscript{1}DFKI - German Research Center for Artificial Intelligence \hspace{0.5cm} \textsuperscript{2}BMW Group \\
{\tt\small firstname.lastname@\string{bmw,dfki\string}.de}
}

\maketitle

\begin{abstract}
	Training a deep neural network is a non-trivial task. Not only the tuning of hyperparameters, but also the gathering and selection of training data, the design of the loss function, and the construction of training schedules is important to get the most out of a model. In this study, we perform a set of experiments all related to these issues. The model for which different training strategies are investigated is the recently presented SDC descriptor network (stacked dilated convolution). It is used to describe images on pixel-level for dense matching tasks. Our work analyzes SDC in more detail, validates some best practices for training deep neural networks, and provides insights into training with multiple domain data.
\end{abstract}

\section{Introduction} \label{sec:intro}
Nowadays, advances in computer vision are dominated by deep learning approaches. The impressive success on various topics and tasks for all kinds of applications catalyzes ever more research in this field.
Though a principled way for learnable representations, classifiers, and regressors is endorsed, our current understanding of deep neural networks lacks behind.
Networks are often handled as black boxes due to the stochastic and iterative nature of back-propagation, the un-interpretable interior of deep and wide architectures, and the increasing number of hyper-parameters.

These facts lead to a conflict for complex, yet safety-critical applications like autonomous driving.
On the one hand, most recent achievements for core components of self-driving cars, like perception or action planning, are enabled by deep learning.
On the other hand, the robustness and reliability of these components remain unexplored which introduces high risk since neither the probability nor the possible maximum harm of wrong decisions is known. 

As a result, we need networks that are more interpretable, more robust (however robustness can be defined), and less self-confident (\ie providing a measure of certainty).

Moreover, part of the success of deep learning is driven by the availability of data. Astonishing results are often obtained only by increasing the amount of training data, using deeper architectures, and thus requiring even more data.
Along with that, the computational effort for training increases likewise, introducing another limiting factor.
While, in principle, there is nothing wrong with using more data, one has to keep in mind that data (labeled or unlabeled) is differently scarce for different domains and applications.
Thus, a working model for one domain might not be transferable to another.
Further, an advanced usage of only very few data is essential to limit the expensive efforts for annotation.
As a conclusion, the available data should be used as efficient as possible to train more accurate and robust models in less time.

In this study, we will focus less on the selection of the architecture, but instead use an existing, shallow model that incorporates an understanding of the given problem into its design \cite{schuster2019sdc}.
Rather, we will investigate effects of training procedures and data in the hope to derive some heuristics that can guide others when training deep neural networks.

Our use case is embedded in the context of environmental perception for automotive applications.
In detail, the network under consideration is the recently presented SDC network \cite{schuster2019sdc} that was designed for image description to aid dense matching tasks, like in optical flow or stereo disparity estimation.
Matching is a mid-level computer vision task that can be used to reconstruct geometry and estimate motions and therefore it builds the foundation for high-level perception and planning tasks which are required for advanced driver assistance systems and autonomous vehicles.

The rest of the paper is structured as follows.
In \cref{sec:background}, we describe some related work and introduce the relevant data sets for our experiments.
The SDC feature description network that we use in our study is explained in \cref{sec:model} along with some deeper analysis.
Our experiments are presented in \cref{sec:study}.
We summarize our results in \cref{sec:conclusion}.

\begin{table*}[t]
	\centering
	\caption{Characteristics of different data sets.} \label{tab:datasets}
	\begin{tabular}{c|ccccccc}
		Data Set & Task & \thead{ Number of\\Sequences} & \thead{Frames per\\Sequence} & \thead{Image\\Size [MP]} & \thead{Color\\Space} & \thead{Synthetic\\Real} & \thead{Automotive\\Context}\Bstrut\Tstrut\\
		\hline
		KITTI \cite{menze2015object} & sf & 200 & 1 & 0.46 & RGB & R & yes \Tstrut\\
		FlyingThings3D \cite{mayer2016large} & sf & 2239 & 10 & 0.52 & RGB & S & no \\
		Driving \cite{mayer2016large} & sf & 1 & 800 & 0.52 & RGB & S & yes \\
		Sintel \cite{butler2012sintel} & mix & 23 & 46 & 0.45 & RGB & S & no \\
		HD1K \cite{kondermann2016hci} & of & 35 & 30 & 2.8 & Gray & R & yes \\
		Middlebury Flow \cite{baker2011database} & of & 8 & 1 & 0.25 & RGB & both & no \\
		Middlebury Stereo \cite{scharstein2002taxonomy} & st & 15 & 1 & 1.1 - 17.4 & RGB & R & no\\
		ETH3D \cite{schops2017multi} & st & 16 & 1 & 0.31 / 0.46 & Gray & R & no \\
	\end{tabular}
\end{table*}

\section{Background} \label{sec:background}

\paragraph{Related Work.}
The importance of training data and schedules for end-to-end optical flow estimation was investigated in \cite{mayer2018what,sun2018models}.
In \cite{mayer2018what}, the usability of synthetic data for transfer learning (in the form of pre-training + fine-tuning) was investigated.
The authors conducted a series of experiments about lighting, data augmentation, displacement statistics, simulation of realistic noise when generating synthetic images, hyperparameter tuning, and the importance of the order when training with multiple data sets. 
The model under review was FlowNet \cite{dosovitskiy2015flownet,ilg2017flownet}.
Advanced training strategies for PWCNet \cite{sun2018pwc} were presented in \cite{sun2018models}.
Here, the focus was to adjust the training process to improve generalization of the network for the Robust Vision Challenge\footnote{\label{foot:rob}\url{www.robustvision.net}}.

The work in this paper conducts a similar empirical study with focus on training strategies for deep neural networks.
Contrary to the previous work, our model of interest is a generic feature description network that is not restricted to the optical flow problem.

\paragraph{Matching Tasks.}
SDC \cite{schuster2019sdc} was presented as a generic feature descriptor that can be used for any dense matching task, \eg stereo, optical flow, or scene flow matching.
Finding image correspondences for these problems is related to different image pairs.
For optical flow (\textit{of}), images are matched in the temporal domain, taken with the same camera.
For stereo matching (\textit{st}), we have two distinct rectified cameras that capture images simultaneously.
A combination of both (\textit{mix}) is possible if a data set provides ground truth for optical flow and stereo disparity.
If the annotations further provide a measure for the change of depth, image correspondences between stereo cameras over time (\textit{cross}, (\textit{cr})) can be established.
A data set that contains labels for \textit{st}, \textit{of}, and \textit{cr} is capable of training full scene flow (\textit{sf}) matching.
In \cref{sec:study}, we will show that these matching tasks have quite different characteristics.

\paragraph{Data Sets.}
As mentioned in the introduction, data is of utmost importance for training. Increasing effort is spent on capturing, labeling, or generation of large data sets for different domains to enable training of deeper and larger models.
Generalization to unseen samples -- and even more to unseen domains -- remains a challenging problem for neural networks. Yet, a tendency to overcome this issue by extensive use of more and diverse data is evident in recent publications \cite{bailer2017cnn,sun2018pwc}.

For many applications it is hard, tedious, or impossible to collect labeled training data (\eg optical flow) even considering manual annotation. Therefore, synthetic data sets are often used for training followed by fine-tuning on the target domain to transfer what was learned.
Advantages of synthetic data generation include large scale and dense, exact ground truth annotations. However, image appearance (even if photo-realistic) might not fit the realistic data, thus increasing the problems of generalization, overfitting, and domain adaption.

One synthetic data set, that is relevant for our work, is FlyingThings3D (FT3D) \cite{mayer2016large} since it is, besideds KITTI \cite{geiger2012kitti,menze2015object}, the only other data set providing full scene flow labels.
Especially the \textit{Driving} subset of FT3D is relevant, because it simulates a traffic scenario.
MPI Sintel \cite{butler2012sintel} is also quite large and provides optical flow and stereo labels, making it a possible candidate for deep training.

Among realistic data sets, KITTI \cite{menze2015object} is the natural choice since it provides scene flow ground truth (though sparse) in an automotive context.
The data of HD1K \cite{kondermann2016hci} is also captured from a stereo camera mounted on a driving vehicle, but it provides only annotations for optical flow correspondences.
The original SDC network was additionally trained on the other data sets that are part of the Robust Vision Challenge\cref{foot:rob} for stereo and optical flow (Middlebury (MB) \cite{scharstein2002taxonomy,baker2011database} and ETH3D \cite{schops2017multi}).
However, the latter three are not suitable for training because they are very limited in size.
An overview of all these data sets is given in \cref{tab:datasets}.

\begin{figure*}[t]
	\centering
	\includegraphics[width=1.0\linewidth]{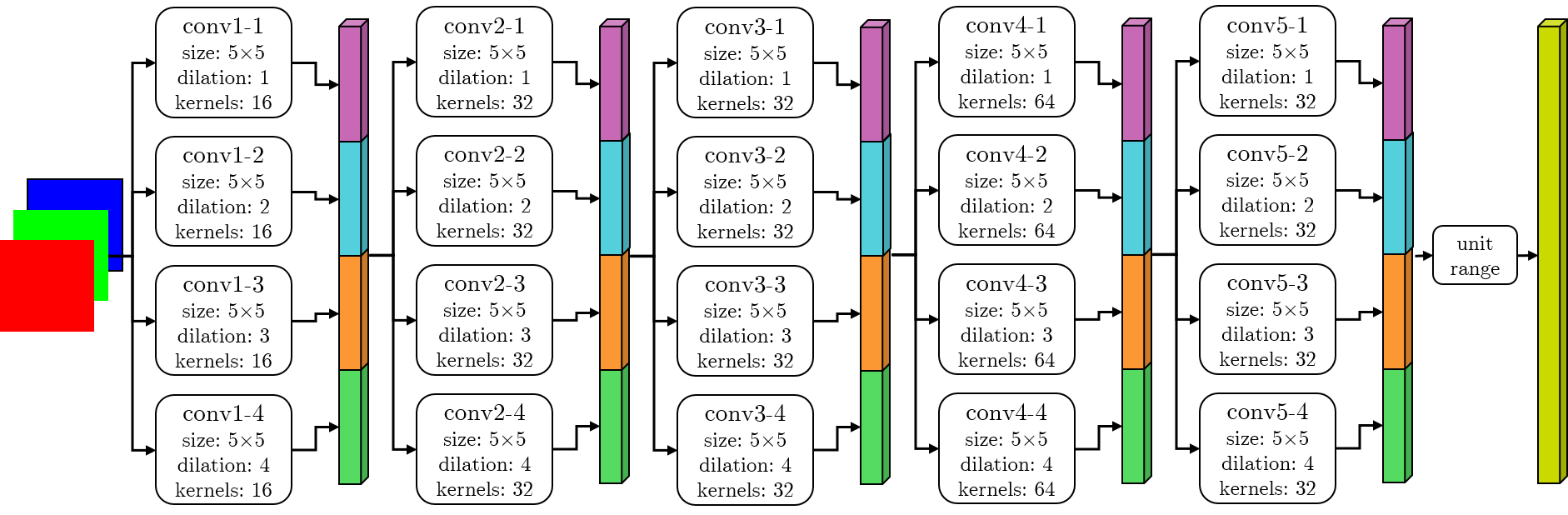}
	\caption{The architecture of the SDC feature descriptor network \cite{schuster2019sdc}.}
	\label{fig:network}
\end{figure*}

\section{SDC Features} \label{sec:model}

\paragraph{Network Architecture.}
The SDC network for feature description \cite{schuster2019sdc} was recently published and demonstrated superior performance over heuristic descriptors like SIFT \cite{lowe1999sift} when applied in state-of-the-art matching algorithms (ELAS \cite{geiger2010elas}, SGM \cite{hirschmuller2008SGM}, CPM \cite{hu2016efficient}, FlowFields++ \cite{schuster2018ffpp}, and SceneFlowFields \cite{schuster2018sceneflowfields}).
SDC was further shown to be more accurate and robust in patch matching compared to other feature networks.
Its properties and the presented experiments make SDC a good candidate for generic feature computation in all kinds of architectures and applications.

The SDC network uses the concept of stacked dilated convolutions which is motivated by the observation, that dilated convolution is equivalent to regular convolution on sub-sampled input data. Therefore, concatenating the output of parallel dilated convolutions is producing a multi-scale feature representation of the input.

The proposed architecture of \cite{schuster2019sdc} consists of five such stacked dilated convolution layers, each with 4 parallel convolutions with $5 \times 5$ kernels and dilation rates $d = 1,2,3,4$. This setup yields a receptive field of $81$ pixels with a dense feature prediction for every input pixel. The complete structure of the SDC network is visualized in \cref{fig:network}.

\paragraph{Training.}
The original SDC model was trained with a mixture of data from KITTI \cite{menze2015object}, Sintel \cite{butler2012sintel}, HD1K \cite{kondermann2016hci}, Middlebury (MB) \cite{scharstein2002taxonomy,baker2011database}, and ETH3D \cite{schops2017multi}.
The ratio of used training patches was 0.5, 0.175, 0.175, 0.05/0.025, and 0.075 respectively, which is in accordance to the variance and scale of the labeled data of each data set as shown in \cref{tab:datasets}.
The optimizer in \cite{schuster2019sdc} is ADAM \cite{kingma2015adam} with a progressive learning rate decay (cf. \cref{fig:disruption}). In \cite{schuster2019sdc}, a triplet training strategy was applied, where a reference image patch along with the corresponding and a non-corresponding patch are sampled randomly. SDC was trained with batches of 32 triplets for 1 million iterations with a thresholded hinge embedding loss \cite{bailer2017cnn}.

\begin{figure}[t]
	\centering
	\begin{overpic}[width=1.0\linewidth]{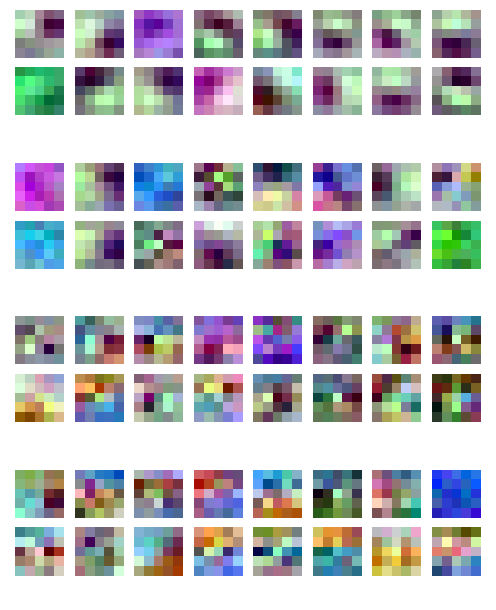}
		\put(40.5,79){\makebox(0,0){\footnotesize Dilation: 1}}
		\put(40.5,53.5){\makebox(0,0){\footnotesize Dilation: 2}}
		\put(40.5,28){\makebox(0,0){\footnotesize Dilation: 3}}
		\put(40.5,3.5){\makebox(0,0){\footnotesize Dilation: 4}}
	\end{overpic}
	\caption{Convolution kernels for the first SDC layer of the SDC feature network \cite{schuster2019sdc}. The color gives the respective sensitivity to the RGB color channels of the input images.}
	\label{fig:kernels}
\end{figure}

\begin{figure}[t]
	\centering
	\begin{subfigure}[c]{0.7\linewidth}
		\includegraphics[width=1.0\linewidth]{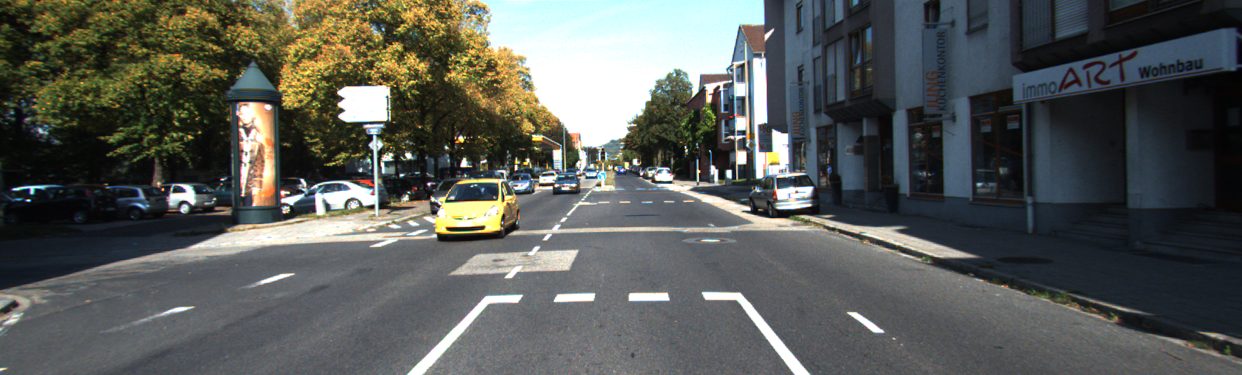}
		\caption{Input Image} \label{fig:features:img}		
	\end{subfigure}\\
	\begin{subfigure}[c]{0.7\linewidth}
		\includegraphics[width=1.0\linewidth]{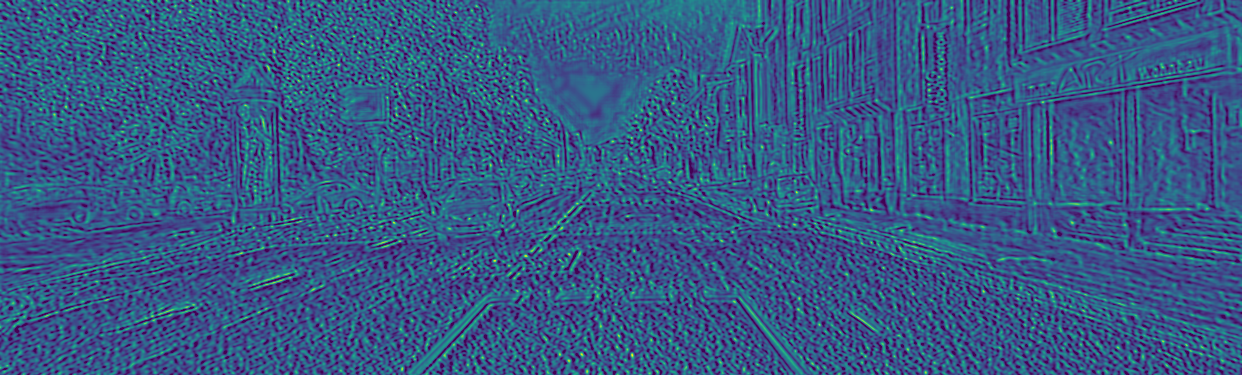}
		\caption{Channel 26} \label{fig:features:d1}		
	\end{subfigure}\\
	\begin{subfigure}[c]{0.7\linewidth}
		\includegraphics[width=1.0\linewidth]{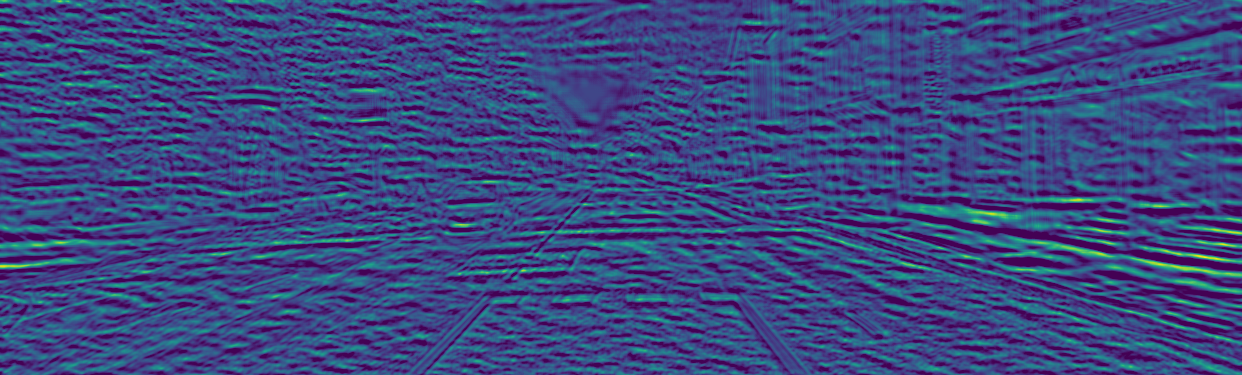}
		\caption{Channel 53} \label{fig:features:d2}		
	\end{subfigure}\\
	\begin{subfigure}[c]{0.7\linewidth}
		\includegraphics[width=1.0\linewidth]{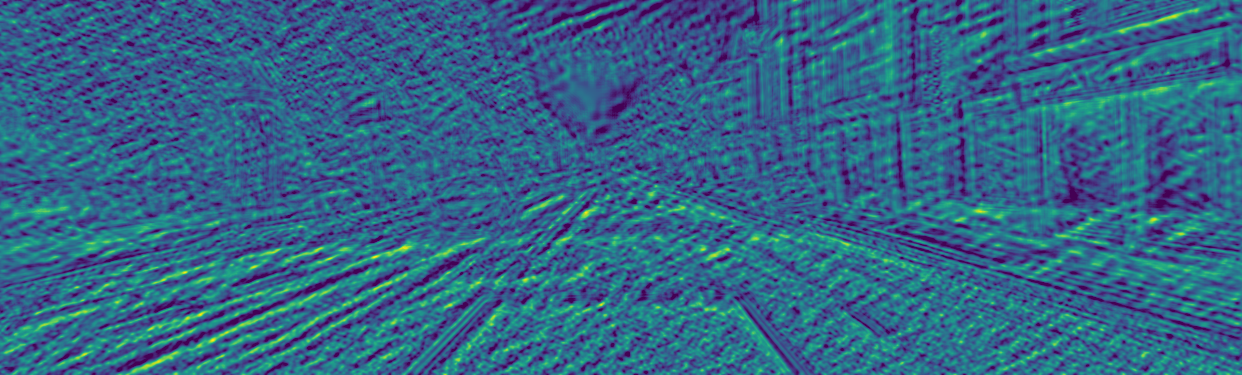}
		\caption{Channel 66} \label{fig:features:d3}		
	\end{subfigure}\\
	\begin{subfigure}[c]{0.7\linewidth}
		\includegraphics[width=1.0\linewidth]{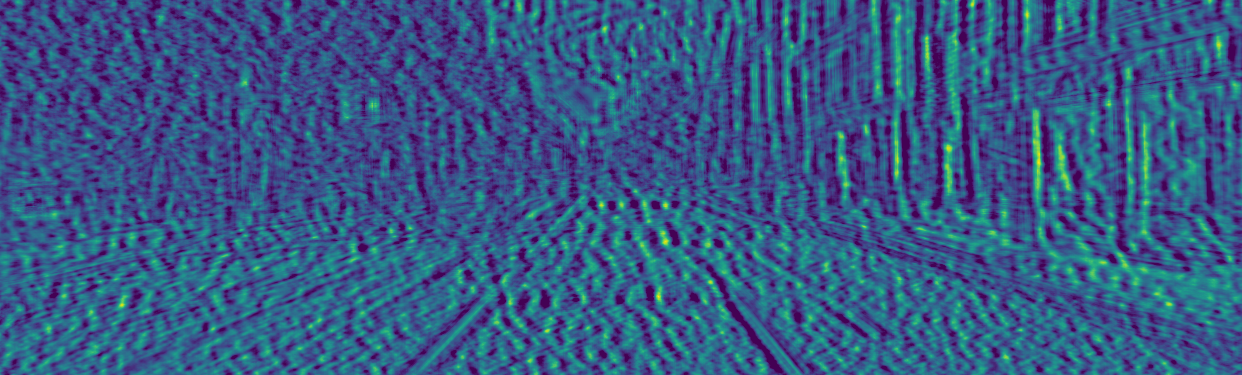}
		\caption{Channel 103} \label{fig:features:d4}		
	\end{subfigure}
	\caption{Some SDC feature channels for the given input image.}
	\label{fig:features}
\end{figure}

\paragraph{Feature Analysis.}
The original training strategy is used for our deeper analysis of SDC features.
First, we visualize the learned kernels of the first SDC layer (see \cref{fig:kernels}). 
The first learned filters with a dilation rate of 1 show a high similarity to two-dimensional second order Gaussian kernels.
For higher dilation rates, the kernels become less intuitive.
There are also some filters that respond to a certain color. 

Next, we present some of the normalized filter responses of the last SDC layer, \ie the final feature representation, for an exemplary image in \cref{fig:features}. 
Different channels for coarse and fine structures can be identified clearly. 
One special observation is, that one feature channel dominates the representation, \ie all values are 1, the maximum. 
Further experiments showed that this dimension is the same for all investigated images on all data sets.
Also interesting is the fact that more than one third of all dimensions does not contribute to the description significantly, \ie the features for these channels are all very close to zero for all data sets.
The amount of "dead channels" decreases for increasing dilation rates (conv5-1: 18, conv5-2: 16, conv5-3: 12, conv5-4: 7).
However, the remaining channels (not 0 and not 1) are all equally important for description according to their variance.

\begin{figure}[t]
	\centering
	\hspace{0.05\linewidth}
	\begin{subfigure}[c]{0.25\linewidth}
		\centering \footnotesize Positive\\Patch
	\end{subfigure}
	\begin{subfigure}[c]{0.25\linewidth}
		\centering \footnotesize Reference\\Patch
	\end{subfigure}
	\begin{subfigure}[c]{0.25\linewidth}
		\centering \footnotesize Negative\\Patch
	\end{subfigure}\\%
		\vspace{1mm}%
	\begin{subfigure}[c]{0.05\linewidth}
		\rotatebox[origin=c]{90}{\footnotesize Occlusion}
	\end{subfigure}
	\begin{subfigure}[c]{0.25\linewidth}
		\includegraphics[width=1.0\linewidth]{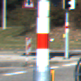}
	\end{subfigure}
	\begin{subfigure}[c]{0.25\linewidth}
		\includegraphics[width=1.0\linewidth]{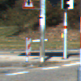}
	\end{subfigure}
	\begin{subfigure}[c]{0.25\linewidth}
		\includegraphics[width=1.0\linewidth]{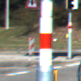}
	\end{subfigure}\\%
		\vspace{1mm}%
	\begin{subfigure}[c]{0.05\linewidth}
		\rotatebox[origin=c]{90}{\footnotesize Boundary}
	\end{subfigure}
	\begin{subfigure}[c]{0.25\linewidth}
		\includegraphics[width=1.0\linewidth]{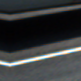}
	\end{subfigure}
	\begin{subfigure}[c]{0.25\linewidth}
		\includegraphics[width=1.0\linewidth]{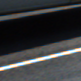}
	\end{subfigure}
	\begin{subfigure}[c]{0.25\linewidth}
		\includegraphics[width=1.0\linewidth]{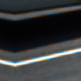}
	\end{subfigure}\\%
		\vspace{1mm}%
	\begin{subfigure}[c]{0.05\linewidth}
		\rotatebox[origin=c]{90}{\footnotesize Homogeneous}
	\end{subfigure}
	\begin{subfigure}[c]{0.25\linewidth}
		\includegraphics[width=1.0\linewidth]{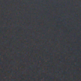}
	\end{subfigure}
	\begin{subfigure}[c]{0.25\linewidth}
		\includegraphics[width=1.0\linewidth]{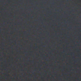}
	\end{subfigure}
	\begin{subfigure}[c]{0.25\linewidth}
		\includegraphics[width=1.0\linewidth]{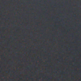}
	\end{subfigure}\\%
		\vspace{1mm}%
	\begin{subfigure}[c]{0.05\linewidth}
		\rotatebox[origin=c]{90}{\footnotesize Vegetation}
	\end{subfigure}
	\begin{subfigure}[c]{0.25\linewidth}
		\includegraphics[width=1.0\linewidth]{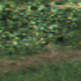}
	\end{subfigure}
	\begin{subfigure}[c]{0.25\linewidth}
		\includegraphics[width=1.0\linewidth]{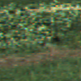}
	\end{subfigure}
	\begin{subfigure}[c]{0.25\linewidth}
		\includegraphics[width=1.0\linewidth]{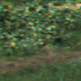}
	\end{subfigure}\\%
		\vspace{1mm}%
	\begin{subfigure}[c]{0.05\linewidth}
		\rotatebox[origin=c]{90}{\footnotesize Object}
	\end{subfigure}
	\begin{subfigure}[c]{0.25\linewidth}
		\includegraphics[width=1.0\linewidth]{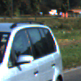}
	\end{subfigure}
	\begin{subfigure}[c]{0.25\linewidth}
		\includegraphics[width=1.0\linewidth]{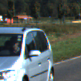}
	\end{subfigure}
	\begin{subfigure}[c]{0.25\linewidth}
		\includegraphics[width=1.0\linewidth]{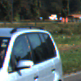}
	\end{subfigure}	
	\caption{Misclassified triplets from the KITTI test split for the original SDC network.}
	\label{fig:failurecases}
\end{figure}

\paragraph{Failure Cases.}
We evaluate SDC \cite{schuster2019sdc} on a test set of patch triplets. A triplet is considered as misclassified, if the feature distance of the corresponding image patches is smaller than the feature distance of non-corresponding patches (cf. \cref{sec:study}).
Some representative, misclassified triplets from the KITTI test set are depicted in \cref{fig:failurecases}.
The failure cases can be clustered into the following categories where one triplet can belong to multiple classes: \textit{Vegetation} (34~\%), \textit{dynamic objects} (29~\%), \textit{occlusions} (18~\%), \textit{boundary regions} (16~\%), \textit{homogeneous patches} (14~\%).
While homogeneous, untextured and occluded regions can only be matched with a wider receptive field (\ie changing the architecture to consider more context knowledge), the issues of dynamic foreground objects and vegetation can be tackled by changing the training schedule as done in the next sections.
The only reliable way to handle image boundaries is to ignore them during feature computation and matching.

\section{Empirical Study} \label{sec:study}
The experiments within this section are split into two groups.
First, we investigate how training can be improved in general.
The second part focuses more on data and topics related to training on multiple domains.
Unless stated otherwise in our experiments, a single data set model is always trained on all available image pairs (\eg KITTI uses all three image pairs of the scene flow).
As major evaluation criterion, the triplet accuracy is used. That is the percentage of properly distinguished patch triplets (corresponding feature distance is smaller than non-corresponding feature distance).

\subsection{Improved SDC Training} \label{sec:study:single}

\begin{figure}[t]
	\centering
	\begin{subfigure}[c]{1\linewidth}
		\includegraphics[width=1\linewidth]{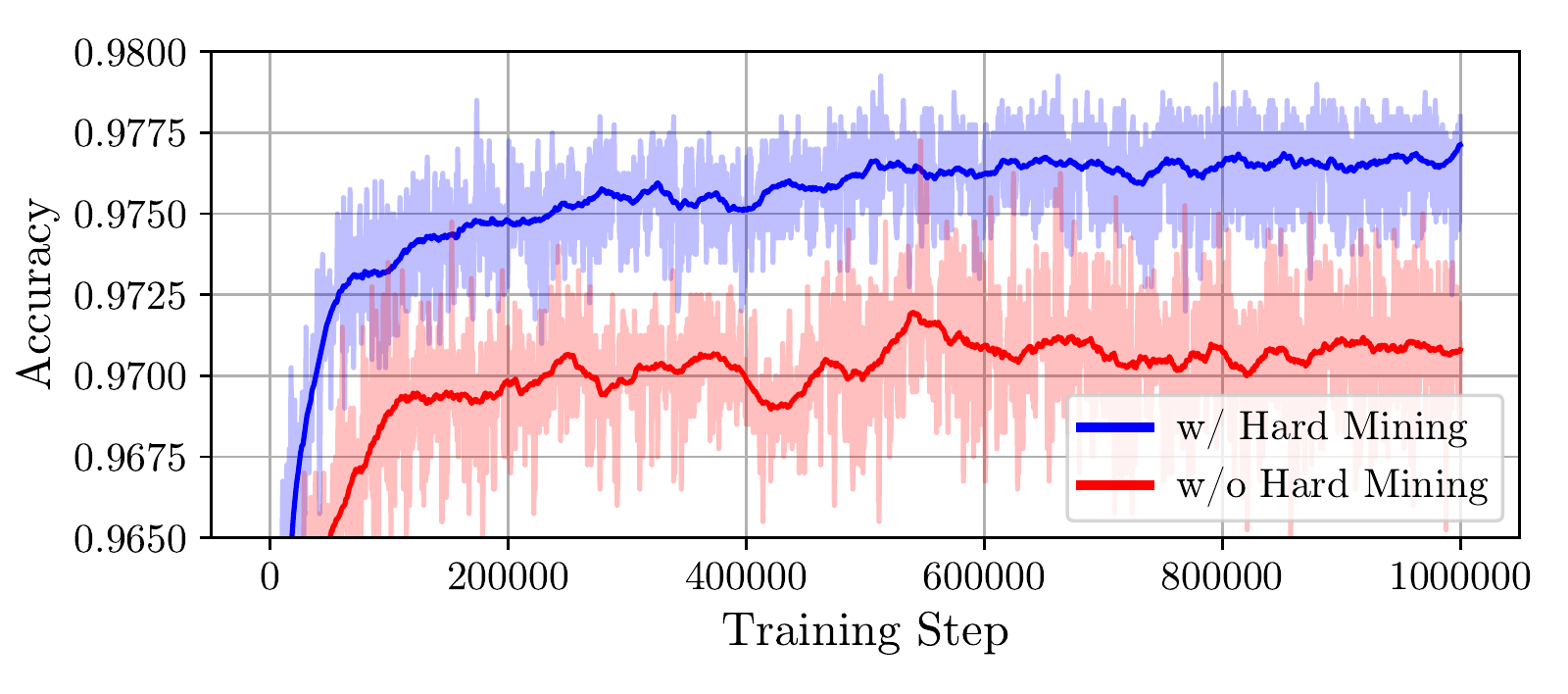}
		\caption{Hard Mining.} \label{fig:training:hardmining}
	\end{subfigure}
	\begin{subfigure}[c]{1\linewidth}
		\includegraphics[width=1\linewidth]{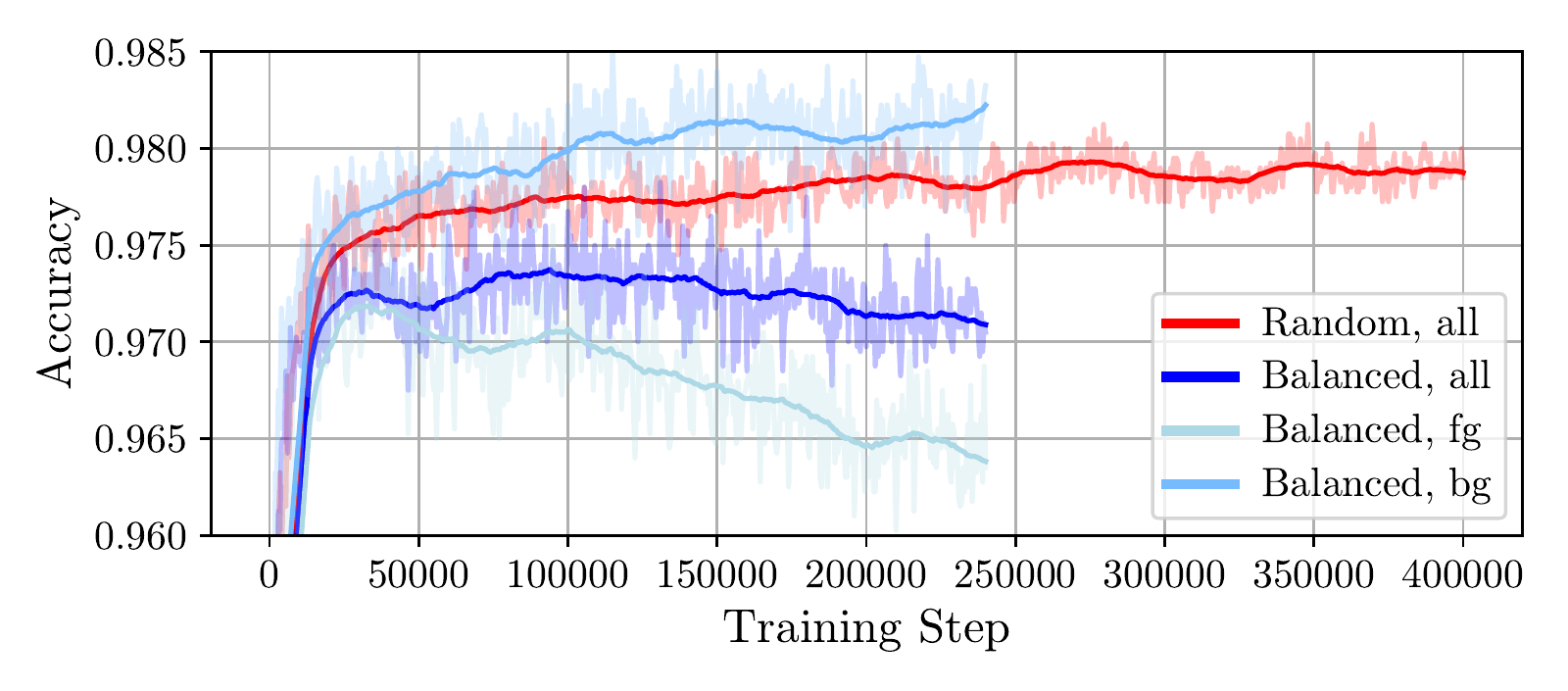}
		\caption{Balanced Region Sampling.} \label{fig:training:balancing}
	\end{subfigure}
	\begin{subfigure}[c]{1\linewidth}
		\includegraphics[width=1\linewidth]{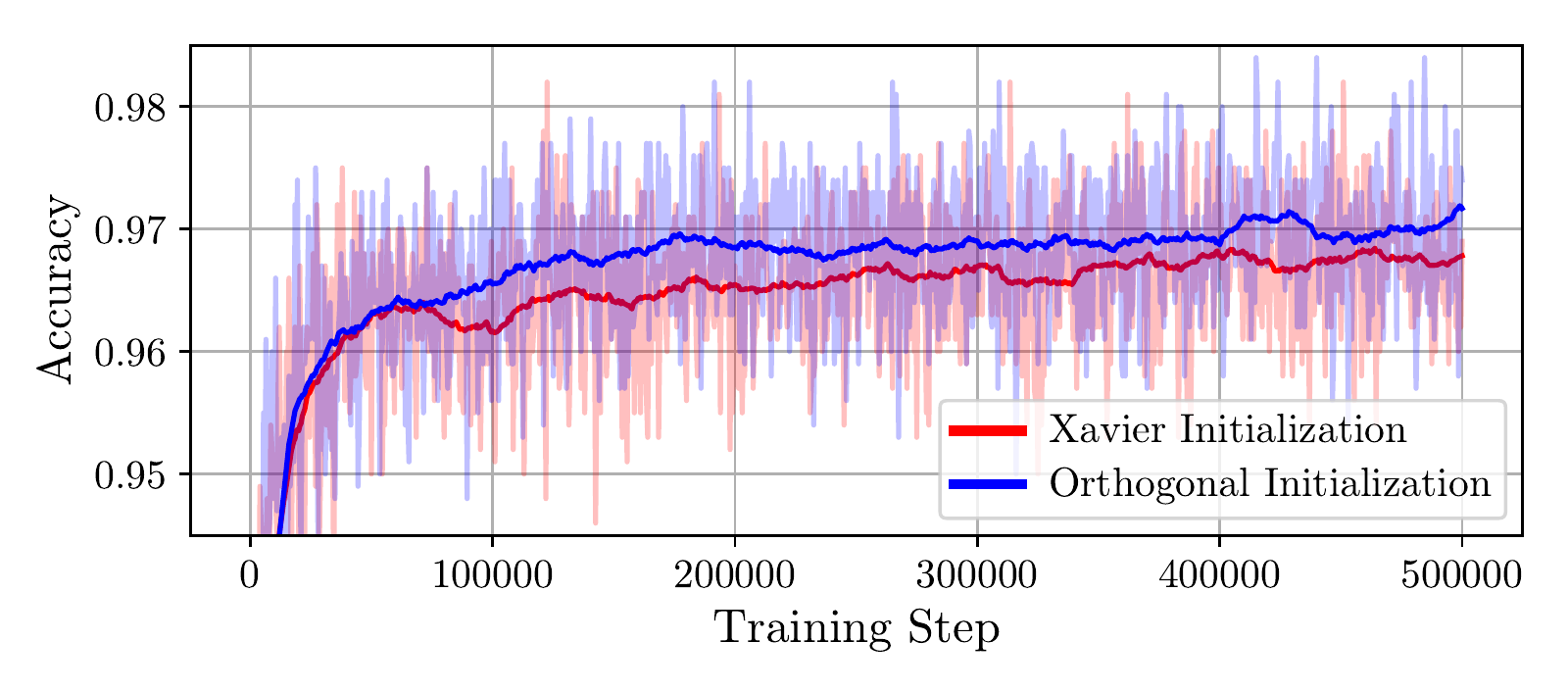}
		\caption{Weight Initialization.} \label{fig:training:initialization}
	\end{subfigure}
	\caption{Comparison of different training setups.}
	\label{fig:training}
\end{figure}

\paragraph{Hard Mining.}
Hard mining is a well documented technique to speed up training and increase the accuracy especially for difficult samples \cite{schroff2015facenet}. It is also helpful when training with imbalanced data \cite{dong2017class}. 
The idea is to ignore samples with a sufficiently accurate prediction during training and focus more on samples with less accurate or wrong predictions. 
In our case, we implement offline hard mining by ignoring triplets with zero loss, \ie positive distance is below a threshold and the negative distance is higher than the margin (cf.~the thresholded hinge embedding loss in \cite{schuster2019sdc}).
The expected behavior of training with hard mining is threefold.
First of all, we expect higher (average) losses since zero losses are neglected.
Secondly, training should be speeded up because higher losses lead to higher gradients in more relevant directions.
Lastly, the predictions for difficult samples should be more accurate.
\Cref{fig:training:hardmining} shows the validation accuracy during training with and without hard mining.
Not only is the training much faster, it also reaches a higher final accuracy.

\paragraph{Region Sampling.}
Foreground objects on KITTI are one of the identified failure categories. In \cite{schuster2019sdc}, the authors argue that this is due to the under-representation of dynamic foreground in the KITTI data set (only about 15~\% of the available ground truth).
Apart from hard mining, we can tackle this issue by manually balancing different image regions during patch sampling.
Since ground truth object segmentation is available for KITTI training images, we can sample our reference patches for training equally often from foreground objects and static background regions.
A comparison between balanced sampling and uniform random sampling is presented in \cref{fig:training:balancing} by plotting the validation accuracy during training on KITTI optical flow data for different image regions (foreground (\textit{fg}) / background (\textit{bg}) / \textit{all}).
It is evident in this diagram that balanced sampling leads to very early over-fitting in foreground regions, thus hindering convergence of the model. 
As a result, not even the foreground regions are similarly well described as with uniform random sampling.

\begin{figure}[t]
	\centering
	\includegraphics[width=1.0\linewidth]{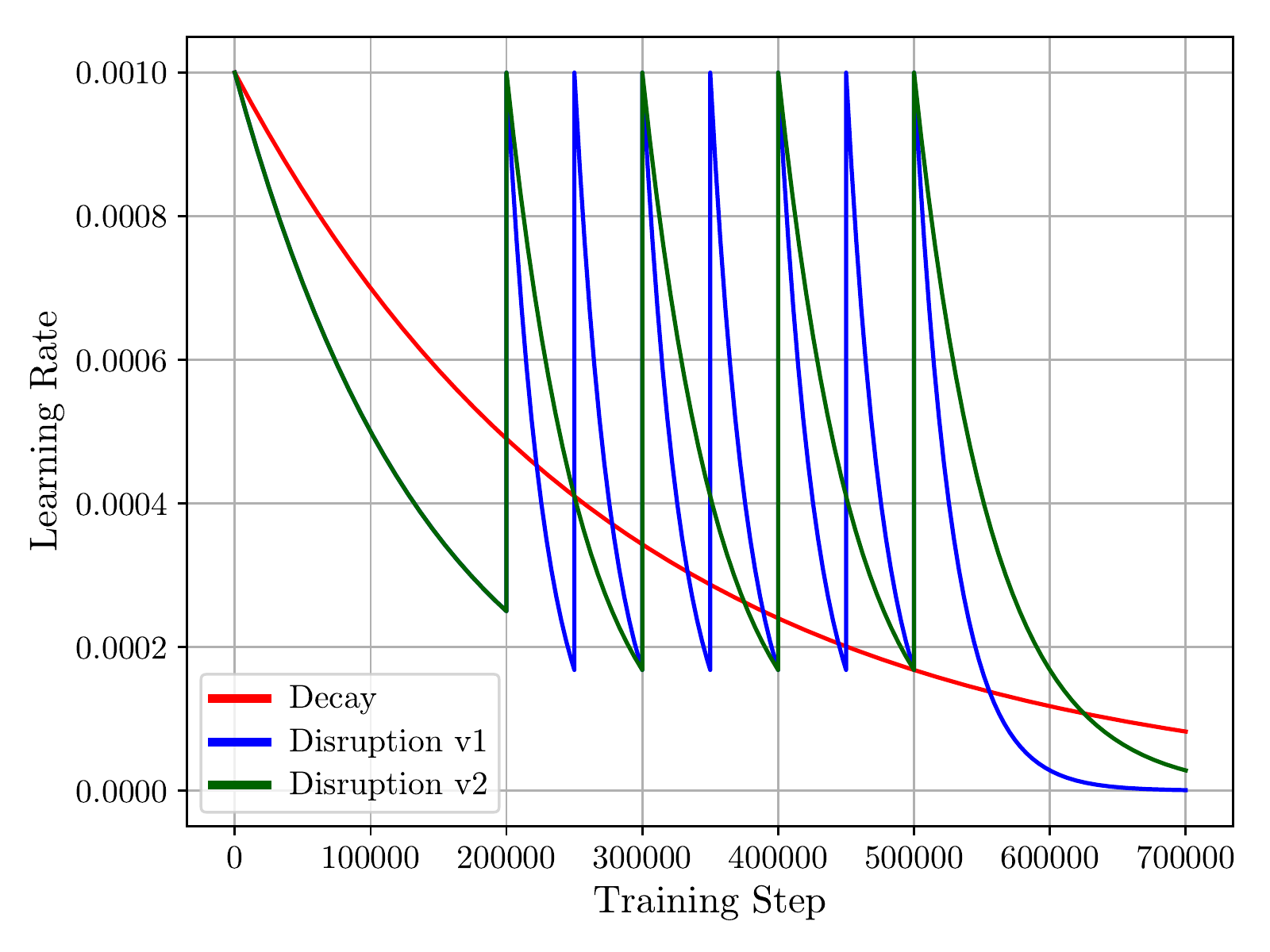}
	\caption{Monotonic decreasing learning rate schedule and two versions for learning rate disruption.}
	\label{fig:disruption}
\end{figure}

\begin{table*}[t]
	\centering
	\caption{Cross evaluation for different domains represented by different data sets. For each evaluation set, the best model trained with a different data set is given in bold.}
	\label{tab:domains}
	\resizebox{1\textwidth}{!}{\begin{tabular}{c|ccccc|ccccc|ccccc|ccc|c|cc|c}
		\multirow{2}{*}{\backslashbox{Train}{Eval}} & \multicolumn{5}{c|}{KITTI} & \multicolumn{5}{c|}{FT3D} & \multicolumn{5}{c|}{Driving} & \multicolumn{3}{c|}{Sintel} & \multicolumn{1}{c|}{HD1K} & \multicolumn{2}{c|}{MB} & \multicolumn{1}{c}{ETH3D}\\
		& sf & mix & cr & fl & st & sf & mix & cr & fl & st & sf & mix & cr & fl & st & mix & fl & st & fl & fl & st & st\Bstrut\\
		\hhline{-|-----|-----|-----|---|-|--|-}
		KITTI \cite{menze2015object} & {\cellcolor{gray!30}} 97.2 & {\cellcolor{gray!30}} 97.7 & {\cellcolor{gray!30}} 97.8 & {\cellcolor{gray!30}} 97.9 & {\cellcolor{gray!30}} 96.2 & 91.4 & 91.5 & 90.1 & 93.5 & 90.1 & \textbf{68.6} & \textbf{70.7} & \textbf{64.5} & 66.7 & \textbf{75.7} & 90.0 & 89.3 & 90.7 & \textbf{98.5} & 98.8 & 90.3 & 95.5\Tstrut\\
		FT3D \cite{mayer2016large} & 73.9 & 76.5 & 74.6 & 76.9 & 73.3 & {\cellcolor{gray!30}} 95.1 & {\cellcolor{gray!30}} 95.5 & {\cellcolor{gray!30}} 93.7 & {\cellcolor{gray!30}} 96.7 & {\cellcolor{gray!30}} 94.4 & 57.4 & 60.8 & 51.7 & 51.8 & 71.5 & \textbf{95.3} & \textbf{92.7} & \textbf{95.3} & 97.5 & 96.8 & 82.2 & \textbf{96.7} \\
		Driving \cite{mayer2016large} & 89.3 & 90.8 & 86.7 & 89.6 & 90.6 & 89.9 & 90.0 & 89.1 & 92.0 & 88.5 & {\cellcolor{gray!30}} 75.2 & {\cellcolor{gray!30}} 75.8 & {\cellcolor{gray!30}} 74.2 & {\cellcolor{gray!30}} 74.6 & {\cellcolor{gray!30}} 76.7 & 88.7 & 89.0 & 88.7 & 97.0 & 99.2 & 85.9 & 92.7\\
		Sintel \cite{butler2012sintel} & \textbf{93.5} & \textbf{94.6} & \textbf{92.3} & \textbf{95.2} & \textbf{92.9} & \textbf{92.7} & \textbf{92.8} & \textbf{91.6} & \textbf{94.2} & \textbf{91.8} & 59.8 & 62.7 & 55.2 & 55.6 & 70.8 & {\cellcolor{gray!30}} 92.3 & {\cellcolor{gray!30}} 92.7 & {\cellcolor{gray!30}} 93.0 & 97.2 & \textbf{99.5} & \textbf{91.4} & 94.2\\
		HD1K \cite{kondermann2016hci} & 91.0 & 92.0 & 90.6 & 92.8 & 90.8 & 91.6 & 91.7 & 90.7 & 93.7 & 90.3 & 66.2 & 68.2 & 64.0 & \textbf{67.0} & 69.7 & 88.7 & 88.0 & 88.0 & {\cellcolor{gray!30}} 99.5 & 99.0 & 89.3 & 95.9\\
	\end{tabular}}
\end{table*}

\paragraph{Initialization.}
The high-dimensional, highly non-linear and non-convex functional together with a stochastic iterative optimization technique makes neural networks sensitive to initialization.
Depending on the activation function, \cite{he2015delving,glorot2010understanding} propose random initialization that considers the scale of the previous layer.
Orthogonal initialization \cite{saxe2013exact} was proposed for the use in linear fully-connected layers.
The authors could also demonstrate positive effects with networks that use non-linear activation and convolutional layers.
We compare a state-of-the-art variance scaling initializer \cite{glorot2010understanding} with orthogonal initialization \cite{saxe2013exact} for the SDC Network in \cref{fig:training:initialization}.
Even for the shallow, fully-convolutional SDC Network with ELU activation \cite{clevert2015fast}, orthogonal initialization speeds up the training by about a factor of 2. The final accuracy is also slightly higher.

\paragraph{Learning Rate Disruption.}
Initialization is important for stochastic processes and so is the learning rate for the optimizer.
Progressively (either in steps or continuous) decreasing learning rates are a best practice to enable convergence to local optima.
However, with monotonically decreasing learning rates, the optimizer can not escape local optima.
A measure to encounter this is learning rate disruption, as used \eg in \cite{sun2018models}.
The idea is to disrupt the learning rate schedule by increasing the learning rate significantly (\eg to the initial value) and then continue with the progressive learning rate decay.
This way, the optimizer can escape from a local optimum (though not necessarily in favor of a better optimum).
We have experimented with this concept when training the SDC network.
\Cref{fig:disruption} shows three alternate learning rate schedules.
The original monotonic decrease used in \cite{schuster2019sdc} and two variants of learning rate disruption with different periods for recovery.
We could observe some signs of overfitting right after the disruption. However, the network did recover quickly but without any significant sign of changing the local optima (neither in a positive nor negative way).

\subsection{Multi Domain Training} \label{sec:study:multi}

\paragraph{Domain Similarity.}
As approximation of the similarity of domains, we train mono-domain networks on a single data set and cross-evaluate them on all data sets. \Cref{tab:domains} shows the evaluation matrix for all trained models on all data sets.
We do not train models on the Middlebury (MB) data sets \cite{scharstein2002taxonomy,baker2011database} or ETH3D \cite{schops2017multi} because of their small size.
For the other data sets, we train with the union of all available image correspondences (the three scene flow image pairs for KITTI, FT3D, and Driving, and optical flow and stereo correspondences for Sintel). 
Training all combinations of data sets and tasks would be infeasible.

We observe that domain transfer is not necessarily symmetric.
More over, the matching task (\ie the type of image correspondences) has influence on the matching performance.
Matching on the Driving data set is particularly difficult. On HD1K \cite{kondermann2016hci}, matching is extremely simple. Probably because the ground truth does not contain any dynamic objects. Performance for all models is similarly high for the Middlebury Flow data  \cite{baker2011database}. Most likely because the displacements are very small.
A model trained on Sintel \cite{butler2012sintel} shows high compatibility with many diverse data sets.

Our overall observation is that domain similarity for matching is mostly defined by the displacement characteristics and camera hardware, and less by the scenario or realism of the data.
The Driving data set for example shows a big discrepancy to KITTI in the cross-evaluation, though both contain traffic scenarios.
Reversely, Sintel shares neither the realism nor the automotive setting with KITTI, but still demonstrates high compatibility.
This observation is in accordance with the results in \cite{mayer2018what} on displacement statistics for optical flow.
We can further confirm this by an additional experiment.
The Driving data set comes with two different focal lengths (15 and 35 mm).
The two subsets do not differ in anything else.
Performing a cross-evaluation with models trained on KITTI and both versions of Driving, there is a significant loss in domain similarity when switching to the 35 mm focal length, which is also further away from the KITTI camera parameters.
Moreover, transfer between Driving with different focal length does also not work very well.


\paragraph{Color.}
Two questions of interest regarding color spaces are 1.) Which color space provides good generalization properties? and 2.) Does color influence domain adaption?
We investigate the first question by training a model on one data set with two different color spaces (RGB and YUV) and evaluating them on the other data sets (each in the respective color space).
In our experiments, there is no clear sign that one of the two color spaces should be preferred over the other in terms of generalization.
Both models perform similarly on all test data sets.
There is also no sign that YUV or RGB color promote the training process.
To answer the second question, we train two models on KITTI, one with the original RGB color and one with gray scale converted images to match the color space of HD1K.
Intuitively, one would assume the gray scale model to perform better when evaluated on a gray scale data set like HD1K.
Contrary, the result of our experiment showed that the color model achieves a higher accuracy on HD1K data compared to the gray scale model.
However, when swapping training and evaluation data, a model trained on HD1K performed better on KITTI if the images were converted to gray scale.

\paragraph{Scale.}
In a similar fashion, the influence of scale spaces was studied.
HD1K images have much higher resolution compared to KITTI (cf. \cref{tab:datasets}), thus the receptive field of the SDC network ($81 \times 81$ pixels) covers a much smaller part of the visible scene; even more so because the field of view (FOV) of the camera device is smaller (69 \degree{} instead of 90 \degree).
Again, the assumption is that shifting the scale for the training domain towards the scale of the target domain, would improve the transfer.
Once more, in contrast to our expectation, a model trained on down-scaled HD1K data did not perform better on KITTI compared to a model trained on full resolution images.
Here, the inverse experiment (KITTI model evaluated on full resolution and down scaled HD1K data) indicates also that images should not be scaled to achieve better domain transfer. 
This might be due to artifacts introduced by the scaling.

Nonetheless, scale is important for detection and matching.
The SDC network is specifically designed to deal with varying scales through the use of parallel convolutions with different dilation rates \cite{schuster2019sdc}.
\Cref{tab:scales} shows some baseline descriptors, the original SDC network, and a multi-scale model, all evaluated on multiple scales of the KITTI data.
The heuristic descriptors (SIFT \cite{lowe1999sift}, DAISY \cite{tola2010daisy}, BRIEF \cite{calonder2010brief}) show an almost quadratic loss in performance when image size decreases, even if they are supposed to be scale invariant.
For increased image scale, they perform better.
Presumably because smaller patches show fewer deformations, or other variations between images.
The implicit multi-scale design of SDC performs extremely well on different scales, with only a small drop in accuracy.
For SDC, the performance drops also when the input is upsampled.
This is not surprising since the dilation rates can only simulate smaller scales.
A model explicitly trained on multi-scale data amplifies the scale invariance even more, showing almost no degradation of the accuracy when the scale changes.

\begin{table}[t]
	\centering
	\caption{Multi-scale behavior for different descriptors.} \label{tab:scales}
	\begin{tabular}{c|cccc}
		Descriptor & $\times 2$ & Original & $\times 0.5$ & $\times 0.25$\Bstrut\Tstrut\\
		\hline
		\hline
		Multi-scale & \textbf{96.60} & \textbf{97.30} & \textbf{96.85} & \textbf{96.60}\Tstrut\Bstrut\\
		\hline
		SDC \cite{schuster2019sdc} & 94.55 & 97.25 & 96.70 & 93.90\Tstrut\Bstrut\\
		\hline
		BRIEF \cite{calonder2010brief} & 95.00 & 94.00 & 90.50 & 82.15\Tstrut\\
		DAISY \cite{tola2010daisy} & 92.80 & 91.25 & 88.15 & 80.80\\
		SIFT \cite{lowe1999sift} & 93.90 & 89.90 & 81.95 & 73.65\\
	\end{tabular}
\end{table}

\paragraph{Normalization.}
Standardization of the input is useful to remove any bias from the data and to scale features into equal range, making them equally important for training.
A common practice is to remove the mean pixel value and to scale them so that all channels have unit variance.
Surprisingly, standardization is not crucial to train the SDC network.
A model trained on normalized images performs as well as a model trained on the original image data.

Anyway, normalization might also be useful to boost transfer learning by adjusting the pixel distribution to better fit the target domain.
This, of course, is only possible if imagery for the target is available at training time.
When training on a single domain, experiments showed that neither normalization nor a distribution shift help to better generalize to unseen domains.
Yet, when training with a mixture of data (as done in the original SDC network), standardization for each training data set separately improves the performance on unseen domains if the test data is also standardized according to its own statistics.
For training on multiple domains, a unified normalization based on the pixel distribution of the entire data works also very well and is favorable if a single, unified model for different domains is required.

\section{Conclusion} \label{sec:conclusion}
SDC is a neural network architecture with favorable properties for feature description. The implicit multi-scale design, emulated by parallel dilated convolutions, leads to superior matching performance and great invariance to changes of scale.
The analysis of the network and its feature representation brought insights on the weaknesses of SDC features which motivated our adjustments of the training schedule.
Proper weight initialization and hard mining in the loss computation improved the accuracy and speeded up training by a factor of about 4.
More balanced region sampling during generation of training data or learning rate disruption could not improve the networks performance.
The evaluation of similarity for different domains gave useful directions to improve the process of domain adaption and the training on multiple data sets.
We did also investigate the influence of color, scale, and normalization.
The excellent scale invariance of SDC was boosted even more by dedicated multi-scale training.
For future work, we are interested in improving feature description to make use of all feature dimensions and to explicitly model a measure of uncertainty or matching likelihood of image points.

{\small
\bibliographystyle{ieee_fullname}
\bibliography{bib}
}

\end{document}